\documentclass{article}

\usepackage{calc}
\usepackage{etoolbox}
\usepackage{bibentry}
\usepackage[flushleft]{threeparttable}

\usepackage{enumitem}
\SetLabelAlign{bibright}{\hss\llap{[#1]}}
\newcounter{mynum}

\usepackage{fullpage}
\usepackage{times}
\usepackage{fancyhdr,graphicx,amsmath,amssymb}
\usepackage[ruled,vlined]{algorithm2e}
\include{pythonlisting}
\usepackage[numbers]{natbib}

\usepackage[preprint]{neurips_2021}
\usepackage[utf8]{inputenc}

\usepackage[T1]{fontenc}
\usepackage[english]{babel}

\usepackage{subfiles,float,subcaption}

\usepackage{amssymb,amsmath, physics, bm}
\usepackage[ruled,vlined]{algorithm2e}
\usepackage{bbm}
\usepackage{graphicx}
\usepackage{flexisym}
\usepackage{stackengine}

\usepackage[utf8]{inputenc} 
\usepackage[T1]{fontenc}    
\usepackage{hyperref}       
\usepackage{url}            
\usepackage{booktabs}       
\usepackage{amsfonts}       
\usepackage{nicefrac}       
\usepackage{microtype}      
\usepackage{xcolor}         

\DeclareMathOperator*{\argmax}{argmax}
\DeclareMathOperator*{\argmin}{argmin}
\DeclareMathOperator{\EX}{\mathbb{E}}

\title{Model-based Offline Imitation Learning with Non-expert Data}

\author{Jeongwon Park\\UCLA\\
   \And Lin Yang\\UCLA}

\begin{document}
 \setcitestyle{numbers}
\maketitle

\begin{abstract}
    Although Behavioral Cloning (BC) in theory suffers compounding errors, its scalability and simplicity still makes it an attractive imitation learning algorithm. In contrast, imitation approaches with adversarial training typically does not share the same problem, but necessitates interactions with the environment. Meanwhile, most imitation learning methods only utilises optimal datasets, which could be significantly more expensive to obtain than its suboptimal counterpart. A question that arises is, can we utilise the suboptimal dataset in a principled manner, which otherwise would have been idle? We propose a scalable model-based offline imitation learning algorithmic framework that leverages datasets collected by both suboptimal and optimal policies, and show that its worst case suboptimality becomes linear in the time horizon with respect to the expert samples. We empirically validate our theoretical results and show that the proposed method \textit{always} outperforms BC in the low data regime on simulated continuous control domains. 
    
\end{abstract}

\section{Introduction}
Imitation Learning(IL) refers to the setting where a policy does not have access to the ground truth reward function but instead has expert demonstrations. It has shown to be an efficient approach to policy optimization that has been extensively tested on real world applications\cite{reductions,firstbc}. Most previous works on IL focuses on settings where it has limited access to expert demonstrations, where the general strategies include i) BC\cite{noisybc,firstbc} when no environment interactions are allowed ii) Interactive IL with an expert\cite{dagger} iii) Adversarial IL\cite{gail,airl,gcl}, by minimizing some form of divergence between the expert and learner's state action distribution. In this paper, we consider an entirely offline setting as in i). 
\\ \\ In many practical real world cases, we make the realistic assumption that the set of logged datasets is pyramidal\cite{pessimportance}: there are an enormous number of bad trajectories, many mediocre trajectories, a few good trajectories, etc. It follows that when imitation learning algorithms only make usage of the optimal trajectories, it disregards the information stored in the suboptimal trajectories. How best is it then, to utilise those suboptimal trajectories? While there were some works that attempts to learn from suboptimal demonstrations\cite{failed,trex,noisybc,suboptimal}, there were limited works that focuses on the complete offline setting.
\\ \\ 
As in i), offline reinforcement learning (RL)\cite{offlinetutorial} is a setting where the agent only has access to a fixed dataset of experience, and where interactions with the environment are not permitted. It is a setting of great interest when considering applications where deploying suboptimal policies might be costly, for example in healthcare and self-driving\cite{firstbc}. Although it is fair to assume that such settings has accumulated a vast amount of behavior dataset over the years, be it optimal or suboptimal, offline RL entails the additional assumption of the availability of a reward function, or that such experiences are labeled with rewards. Here, we consider a slightly different setting, where we replace the necessity of reward labels with that of a modest amount of expert demonstrations. That is, the setting where the behavior dataset does not include any reward labels, but instead is composed of both suboptimal \textit{and} (a lot less)optimal trajectories. Our motivations are akin to that of inverse RL\cite{maxent,ng,gcl}, and we consider its offline analogue where reward functions are difficult to engineer\cite{ng} but it is relatively easier to provide expert demonstrations instead. 
\\ \\ 
We also formulate connections with imitation learning and offline RL and present an imitation learning algorithm where, by utilising suboptimal demonstrations, its performance is less dependent on the amount of optimal demonstrations. We approach this problem by utilising a learned model trained on the behavior dataset. Previously, imitation learning with offline model estimation\cite{modelbased} has not been studied thoroughly, as in the infinite state action space regime, a model trained with limited and narrow coverage datasets, such as those that are exclusively comprised of expert demonstrations, is almost  guaranteed empirically to be too incorrect for reliable usage\cite{strictlybatch, mopo}. We then argue that, if we cannot do substantially better with respect to the algorithm, \textit{why don't we change the setting itself}? Namely, we change it to the one presented in the paragraph above, and utilise suboptimal data that is potentially idle. From this stems our motivation; in such instances, can we statistically and empirically demonstrate that, by solely interacting with an estimated transition from such idle data, that we can do better than BC?  
\\
\\
 Under whats hopefully a promising setting, our primary contribution is a general algorithmic framework for offline imitation where a policy is trained adversarially with a discriminator in an estimated model, which is trained a priori. We first demonstrate its statistical benefits over behavioral cloning, and its practical instantiation that outperforms BC on \textit{all} our evaluations. Indeed, offline estimation and usage of a model comes with its own set of difficulties\cite{morel,mopo,mbpo}, as the model only has access to data with limited support and cannot improve its accuracy using additional experience. It is also well known that a policy optimizing in such a model could learn to exploit where it is inaccurate, which degrades the evaluated performance\cite{mbpo,morel}. We provide justifiable remedies that are motivated by recent works in model based offline RL\cite{mopo,morel,combo}, where our algorithm also estimates and penalizes model uncertainty.

\section{Preliminaries}

We consider a $\gamma$ discounted infinite horizon Markov Decision Process (MDP) $M = (\mathcal{S}, \mathcal{A}, T, r, d_0, \gamma)$ where $\mathcal{S}$ and $\mathcal{A}$ denotes the state and action space respectively, $T(s'|s,a)$ the transition dynamics, $r(s,a)$ the reward function, $d_0(s)$ the initial state  distribution, and $\gamma \in (0,1)$ the discount factor. We denote the policy $\pi(a|s)$ as a state conditional action distribution, and $d^\pi_{T}(s)$ as the state distribution induced by $\pi$ in the dynamics $T$, where $p^{\pi}_{T}$ as its state action analogue. We denote $T_t, T_l \in \mathbb{T}$, as the true dynamics and the learned dynamics, respectively in the set of transitions 
$\mathbb{T}$
 \\ \\
 We denote the expert policy as  $\pi^*$, the learner policy as $\hat{\pi}$, and the behavior policy as $\pi^\mu$. In the imitation learning setting, the goal is to find a policy $\hat{\pi}\in \Pi$ that minimizes the difference in expected discounted sum of rewards with that of the expert policy $V^{\pi^*} - V^{\hat{\pi}}$, where we consider $V^{\pi^*}$ to be fixed. In our analysis, $\Pi$ and $\mathbb{T}$ are discrete sets of policies and transitions respectively and we assume a realizable setting for the policy, where $\Pi$ is rich enough for $\pi^* \in \Pi$. 

We follow the conventions in \cite{rltheory}, where for any two policies $\pi$, $\tilde{\pi}$ and dynamics $T$, $\tilde{T}$ we define a discriminating function  $f_{\pi,\tilde{\pi}}^{T, \tilde{T}}$: 
$$f_{\pi,\tilde{\pi}}^{T, \tilde{T}} = \argmax_{f: ||f||_{\infty}\leq 1 } \EX_{s \sim d^{\pi}_{T}}[f(s)] -   \EX_{s \sim d^{\tilde{\pi}}_{\tilde{T}}}[f(s)] $$
and the set of discriminating functions: 
$F = \{f_{\pi,\tilde{\pi}}^{T, \tilde{T}} : \pi,\tilde{\pi} \in \Pi\ , T \in \mathbb{T}, \pi \neq \tilde{\pi} \}$

Note that each $f \in F$ defines an Integral Probability Metric (IPM) of the Total Variation (TV) distance for the two distributions. We also define the data coverage of $p^{\pi^\mu}_{T_t}(s,a)$, $C := \sup_{\pi \in \Pi, s \in S, a \in A}\frac{\nu^\pi_{T_l}(s,a)}{p^{\pi^\mu}_{T_t}(s,a)}$, where $\nu^\pi_{T_l}(s,a)$ denotes an admissible state action density by any $\pi \in \Pi$ in the transition $T_l$, and, $p^{\pi^\mu}_{T_t}(s,a)$ denotes the state action distribution of the behavior policy. Unless otherwise stated, we assume $C$ is bounded. 
We define the model training loss taken with respect to the behavior distribution, $\epsilon_T=\EX_{s,a \sim p^{\pi^\mu}(s,a)}[||
T_t(s'|a,s) - T_l(s'|a,s)||_1] $, and the expected distance of any two policies $\pi$, $\pi^'$ where the expectation is taken with respect to the state vistation distribution of $\pi$ as $\epsilon_{\pi,\pi^'} = \EX_{s \sim d^{\pi}}[||\pi(a|s) - \pi^'(a|s)||_1]$. For simplicity, we assume the existence of an optimization oracle for all optimization problems that follow. We also assume that  $p^{\pi}_{T}(s,a)$ is computable when $\pi$ and $T$ is given. Note that in general we can get arbitrarily close to the true distribution by sacrificing more compute. \\
\textbf{Data Assumption} In our work we assume access to a finite collection of $M$ optimal state action pairs and $N$ suboptimal state action pairs that are generated by the expert and the behavior policy respectively, when rolling out their policy on the true dynamics, i.e $\{s_i^*\, a_i^*\}_{i=1..M} \sim p^{\pi^*}_{T_t}$, $\{s_i^\mu\, a_i^\mu\}_{i=1..N} \sim p^{\pi^\mu}_{T_t}$. Although we focus on instances where $N >> M$, our work is not restricted on that setting. 
\label{gen_inst}

\section{Existing Algorithms}
\label{headings}

\paragraph{BC}
One of the more traditional approaches to imitation learning, behavioral cloning\cite{firstbc} trains a function approximator to replicate the expert under the state distribution induced by the expert. A well known result is that the worst case expected suboptimality for any BC learner scales quadratically in the time horizon with respect to the training error\cite{errorbound,reductions}. We further elucidate by presenting a general bound for the difference in expected returns for any two policies $\pi$ and $\hat{\pi}$. \\ \\
\textbf{Lemma 1}: $$V^{\pi} - V^{\hat{\pi}} \leq \frac{1}{(1-\gamma)^2}\epsilon_{\pi,\hat{\pi}}$$
Proof and further details in Appendix. When $\pi$ and $\hat{\pi}$ are $\pi^*$ and $\pi^{BC}$ respectively, the bound reduces to the well studied error bounds of behavioral cloning under the infinite horizon setting\cite{errorbound}. The difficulty here is that behavioral cloning does not directly minimize the difference in state visitation densities, $\epsilon_{d^{\pi,\hat{\pi}}} := ||d^{\pi}_{T_t}(s) - d^{\hat{\pi}}_{T_t}(s)||_1$, between the learner and the expert. As a result, the error from $\epsilon_{\pi,\hat{\pi}}$ is summed over all timesteps when it propagates to $\epsilon_{d^{\pi,\hat{\pi}}}$, which is also known as covariate shift in the literature\cite{reductions}. Furthermore, $\epsilon_{d^{\pi,\hat{\pi}}}$ is also summed over all timesteps when it propagates to $\epsilon_{V}:= V^{\pi} - V^{\hat{\pi}}$.

$$ \epsilon_{d^{\pi,\hat{\pi}}} \leq \mathcal{O}(\frac{1}{1-\gamma}\epsilon_{\pi,\hat{\pi}}) \;\;\;\;\;\;\; \epsilon_{V} \leq \mathcal{O}(\frac{1}{1-\gamma}\epsilon_{d^{\pi,\hat{\pi}}}) $$
In other words, the error's compounding nature can be attributed to the inability to minimize $\epsilon_{d^{\pi,\hat{\pi}}}$ directly. We can then deduce that if there were indeed a measure of minimizing $\epsilon_{d^{\pi,\hat{\pi}}}$, we would obtain a $\mathcal{O}(\frac{1}{1-\gamma})$ factor improvement in $\epsilon_{V}$. 

\paragraph{Adversarial Imitation}
As shown previously, one gets an unfavorable guarantee on BC when $\epsilon_{\pi,\hat{\pi}}$ is large, which generally occurs when $M$ is small. We also showed that one can reduce the expected worst case error by a factor of $\frac{1}{1-\gamma}$ when the difference in the state marginals are directly minimized. In this section we point out the successes in recent adversarial IL methods\cite{gail,airl,gcl,gametheoretic,gan} and remind that its success over behavioral cloning, particularly in the low-data regime, can simply be attributed to \textit{its access to more information}. While seemingly an obvious statement, the imitation learning literature seems to have overlooked this notion\cite{divergence,errorbound}. Precisely, this additional information comes by the form of $T_t$, which grants them the ability to minimize $\epsilon_{d^{\pi,\hat{\pi}}}$ directly. This is in contrast to the hypothesis presented in \cite{divergence}, where they assign credit to particular differences in formulations of loss functions. 
\\ \\ 
Information theoretically, when the algorithm has access to a true model of its MDP, i.e knowledge of $T_t$, it can compute $d^{\hat{\pi}}_{T_t}(s)$. Subsequently, a rather straightforward routine to compute and minimize $\epsilon_{d^{\pi,\hat{\pi}}}$ is a min-max optimization procedure. \\ \\ 
\textbf{Theorem 2:}\textit{ Let} $$\hat{\pi} = 
\argmin_{\pi \in \Pi}\max_{f \in F} \frac{1}{M}\sum_{m=1}^{{M}} 
f(s^*,a^*) - \EX_{s,a\sim  p^{\pi}_{T_t}(s,a)}[f(s,a)]$$ 
\textit{Then with probability at least} $1-\delta$, 
$$V^{\pi^*} - V^{\hat{\pi}} \leq \frac{2}{1-\gamma}\epsilon_{s}$$
where $\epsilon_s = \sqrt{\frac{ln(\frac{|F|}{\delta})}{M}}$. Proof and further details in Appendix.\cite{rltheory}
\\

Quite simplistically, $\hat{\pi}$ in the algorithm  above is directly minimizing $||p^{\pi^*}_{T_t}(s,a) -p^{\pi}_{T_t}(s,a) ||_{TV}$, albeit in its approximated form. As a result, there is only one stage of error propagation from the policy's objective to $\epsilon_{V}$. Note that the algorithm shares the same essential qualities of most adversarial IL methods\cite{gail,airl,gcl}, \textit{in that additional data are generated} via interactions, after which min-max optimization procedure follows. Essentially, the reduction of dependency on $M$ comes at the cost of the assumption of a known $T_t$, as opposed to the BC setting, where the algorithm only has access to a limited, and usually expensive, samples of expert data. We can then deduce that BC without any new information is a fundamentally limited setting, especially in the low data regime. The recently constructed lower bound on BC constitutes confirmatory evidence of this statement, where the lower bound also scales quadratic in the horizon\cite{imilimits}. However note that such online samples are typically considered expensive and even the most sophisticated algorithms specified for sample efficiency\cite{primal, sampleefficient} generally requires orders of magnitude more number of interactions than $M$.     

\section{Model-based Offline Imitation Learning}
In previous sections we briefly analysed the properties of BC and adversarial IL methods. We concluded that the latter comes with better theoretical guarantees, albeit at the cost of interacting with the environment, or access to a true model. However, BC is still widely used due to its simplicity and scalability\cite{bclimitation}. An immediate question that arises is whether existing and tested methods as the ones discussed can be repurposed for sample efficient learning in a complete offline setting, while conserving the generality and scalability of BC. 

\label{others}
\subsection{Algorithm 1}
We first begin with a simple extension of Theorem 2, namely how its bound differs when the rollouts from the learner is generated from a learned model $T_l$ instead of $T_t$. In both Algorithm 1 and Algorithm 2, a model $T_l$ is trained at initialization from samples $s,a \sim p^{\pi^{\mu}}_{T_t}$. Observing that the expectation in the objective of $\pi$ is taken over $ p^{\pi}_{T_l}(s,a)$ instead of $p^{\pi}_{T_t}(s,a)$, we then obtain the following sample complexity of this modified algorithm. 
\\ \\ 
\textbf{Theorem 3:} \textit{Let} $$\hat{\pi} = 
\argmin_{\pi \in \Pi}\max_{f \in F} \frac{1}{M}\sum_{m=1}^{{M}} 
f(s^*,a^*) - \EX_{s,a\sim  p^{\pi}_{T_l}(s,a)}[f(s,a)]$$
\textit{Then with probability at least} $1-\delta$,
$$V^{\pi^*} - V^{\hat{\pi}} \leq \frac{1}{1-\gamma}(2\epsilon_s + \frac{1}{1-\gamma}2C\epsilon_T ) $$
where $\epsilon_s = \sqrt{\frac{ln(\frac{|F|}{\delta})}{M}}$. Proof and further details in Appendix. 
\\
As in section 3, $\epsilon_s$ is the sampling error that has a $\sqrt{1/M}$ dependency on the number of expert samples. It is precisely the term that we aim to reduce the suboptimality's dependence on. $\epsilon_T$ denotes the training loss of the model. $C$ is the importance weight which serves as a bias correction term. Given a model $T_l$, it accounts for the distributional shift that occurs between the training and evaluation stage, since the distribution in which $T_l$ was trained on,  $p^{\pi^\mu}_{T_t}(s,a)$, differs from the distributions on which it is evaluated. If our assumption on the bounded $C$ holds, both of the error terms are guaranteed to converge to 0 in the limit of infinite samples. However its scaling factor $O(\frac{1}{(1-\gamma)^2})$ implies the coverage of the behavior dataset plays an  essential role in algorithm 1's performance, more so than the expert dataset size. 

Note that Algorithm 1's error, unlike that of BC, incur no dependence on $\epsilon_{\pi^*,\hat{\pi}}$, and only has a $O(\frac{1}{1-\gamma})$ dependency on $M$, albeit at the cost of introducing a model error term $C\epsilon_T$. Intuitively, we 'shift' the quadratic dependency on $M$ obtained in behavioral cloning to the model training loss $\epsilon_T$ in Algorithm 1. Why would this be favorable? Consider the definition of the error term. As an empirical risk minimization (ERM) objective, standard learning theoretical bounds shows that  $\epsilon_T$ is inversely related to the number of $s,a$ pairs from $p^{\pi^\mu}_{T_t}(s,a)$,  However, note that the algorithm is agnostic as to what policy $\pi^\mu(a|s)$ is. In other words, $p^{\pi^\mu}_{T_t}(s,a)$ is the \textit{distribution that is obtained by any behavior policy}, including suboptimal policies. When considering instances where it is much more easier to obtain suboptimal data, our analysis implies that provided sufficient coverage of $p^{\pi^\mu}_{T_t}(s,a)$, such cases do not need as much samples from the expert, since we are 'reallocating' the data dependence from that of the expert to that of any policy.

\subsection{Algorithm 2}
Perhaps in a more practical setting, we might want a scalable alternative loss function that can easily be added to the BC framework along with the statistical benefits. In other words, we search for an algorithm that has similar properties to Algorithm 1, in the sense that the dependency on $M$ is reduced, but can be easily integrated with a BC loss, in the following form: $$ \text{loss}(\pi) = \epsilon_{\pi^*,\pi} + \text{loss}_{auxillary}
$$ 
 This is motivated by the fact that relying solely on $f$ for policy optimization might not be ideal when given a finite dataset, where it is easy to overfit to the discriminator\cite{airl,gcl}. We are able to achieve this by modifying the discriminator function. Algorithm 1 focuses on the setting where $f$ takes in both states and actions as inputs. In this subsection, we study its variant, Algorithm 2, where $f$ instead only discriminates on the state space, while the learner $\hat{\pi}$ also directly minimizes a behavioral cloning loss $\frac{1}{M}\sum_{m=1}^{{M}}||\pi^*(a|s^*) - \pi(a|s^*)||_1 $, such that its final objective becomes a two-way trade off problem $$\text{loss}(\pi) = \frac{1}{M}\sum_{m=1}^{{M}}||\pi^*(a|s^*) - \pi(a|s^*)||_1 - \EX_{s \sim  d^{\pi}_{T_l}}[\hat{f}(s)]\ $$ where $\hat{f} = \argmax_{f \in F} \frac{1}{M}\sum_{m=1}^{{M}} 
f(s^*) - \EX_{s \sim  d^{\pi}_{T_l}}[f(s)]$. We then again bound the algorithm's suboptimality using terms of interest.

\textbf{Theorem 4}:\textit{ Let }
$$\hat{\pi} = \argmin_{\pi \in \Pi}\max_{f \in F} \frac{1}{M}\sum_{m=1}^{{M}} 
f(s^*) - \EX_{s \sim  d^{\pi}_{T_l}}[f(s)] +\frac{1}{M}\sum_{m=1}^{{M}}||\pi^*(a|s^*) - \pi(a|s^*)||_1  $$
\textit{Then with probability at least} $1-\delta$, 
$$V^{\pi^*} - V^{\hat{\pi}} \leq \frac{1}{1-\gamma}(2\epsilon_s + \frac{1}{1-\gamma}2C\epsilon_T +
\epsilon_{\pi})$$
where $\epsilon_s = \sqrt{\frac{ln(\frac{|F|}{\delta})}{M}}$ and $\epsilon_{\pi} = \sqrt{\frac{ln(\frac{1}{\delta})}{M}}$. Proof and further details in Appendix. 
\\
 As in Algorithm 1, both $\epsilon_T$ and $\epsilon_s$ approaches 0 in the limit of infinite samples from the behavior distribution $p^{\pi^{\mu}}_{T_t}$ with high probability. The third term is the sampling error of $\frac{1}{M}\sum_{m=1}^{{M}}||\pi^*(a|s^*) - \pi(a|s^*)||_1$ from the true expectation  $\epsilon_{\pi^*,\pi}$, which also approaches 0 as $M \rightarrow{\infty}$. Thus the derived bound indeed becomes a valid sample complexity of Algorithm 2. As in Algorithm 1, the dependence on $M$ is linear in the time horizon, at the cost of introducing the model error $C\epsilon_T$.  
\\
\\
In the previous section we discussed that the error in the conditional action distribution $\epsilon_{\pi^*,\hat{\pi}}$ of two policies $\pi^*$ and $\hat{\pi}$ adds linearly each consecutive timestep over the time horizon when it propagates to $||d^{\pi^*}_{T_t}(s) - d^{\hat{\pi}}_{T_t}(s)||_1$, and a tighter bound over $V^{\pi^*} - V^{\hat{\pi}}$ necessitates the direct minimization of the error in state densities. A rather simplistic view of Algorithm 2 is that it is minimizing its upper bound (approximated by samples) under its IPM form, $\epsilon_{\pi^*,\hat{\pi}} = 
||d^{\pi^*}_{T_t}(s) - d^{\hat{\pi}}_{T_t}(s)||_1
\leq ||d^{\pi^*}_{T_t}(s) - d^{\hat{\pi}}_{T_l}(s)||_1 + \frac{1}{1-\gamma}C\epsilon_T 
 \cong \frac{1}{M}\sum_{m=1}^{{M}} 
\hat{f}(s^*) - \EX_{s \sim  d^{\hat{\pi}}_{T_l}}[\hat{f}(s)]+\frac{1}{1-\gamma}C\epsilon_T$ (further details in Appendix), where the bound all consists of terms of minimization. From this perspective we can attribute Algorithm 2's sample efficiency over BC by its ability to minimize $\epsilon_{d^{\pi,\hat{\pi}}}$ (almost) directly, which mitigates covariate shift. We also argue that despite the fact that our analysis focuses on a linear loss function for the discriminator $\hat{f}$,  a practical algorithm we propose is quite general.  We do not injure its essence by say, using a different loss function. Rather quite simplistically, its crux is as follows; first train a model, after which, at each iteration, a discriminator and a policy is trained adversarially with $\{s_i^*\}_{i=1..M} \sim d^{\pi^*}_{T_t}$ and $\{\hat{s_i}\}_{i=1..\hat{M}} \sim d^{\hat{\pi}}_{T_l}$, where $\hat{M}$ could be arbitrarily large since we have access to $T_l$. In addition, $\hat{\pi}$ has an additional behavioral cloning loss term to minimize. Full algorithm is detailed in \textbf{Algorithm 2} in Appendix. 
 
\subsection{The \textbf{C} term and its repercussions }

Recall the second term from Theorem 4, $\frac{1}{1-\gamma}2C\epsilon_T$. When training $T_l$ with ERM, the training loss minimized is $\epsilon_T = \EX_{s,a \sim p^{\pi^\mu}_{T_t}(s,a)}[||
T_t(s'|a,s) - T_l(s'|a,s)||_1] $. where the training distribution is limited only to $p^{\pi^\mu}_{T_t}(s,a)$. However when we bound the difference of the witness functions $f$ induced by the the model error(detail in Appendix) $$\EX_{s \sim d^{\hat{\pi}}_{T_t}(s)}[\hat{f}(s)]  - \EX_{s \sim d^{\hat{\pi}}_{T_l}(s)}[\hat{f}(s)]    \leq f_{max}\frac{1}{1-\gamma}\EX_{s,a \sim p^{\hat{\pi}}_{T_l}(s,a)}[||
T_t(s'|a,s) - T_l(s'|a,s)||_1]$$ 
we observe a different distribution  $p^{\hat{\pi}}_{T_l}(s,a)$ at the expectation instead of $p^{\pi^\mu}_{T_t}(s,a)$, i.e a different train and test set for $T_l$. A conventional approach to such case is introducing a concentration coefficient $C$ that is assumed to be bounded uniformly for all $\pi \in \Pi$, $s,a \in S,A $, such that $\EX_{s,a \sim p^{\pi}_{T_l}(s,a)}[||
T_t(s'|a,s) - T_l(s'|a,s)||_1] \leq C\epsilon_T$, ensuring the convergence properties during finite sample analysis, a form of insurance against any distribution shift if you will. Such simple algebraic trick, however, neglects what potentially is a 'make or break' in an offline learning algorithm in a practical setting\cite{offlinetutorial,mopo,morel}. It has been empirically observed\cite{offlinetutorial} that a combination of infinite state-action space and function approximation likely causes divergence while training, due to the policy exploiting the errors in the model, which is expected to be large with such bias in the distributions. 
\\ 
\\
As alluded before, our analysis elucidates this notion, as $C$ compounds at a quadratic rate, which means for very large $C$, the second error term from Theorem 4 will dominate the other terms and the bound will become very loose in the worst case. By its definition, this occurs when \textit{any} $\pi \in \Pi$ admits $s, a$ such that $p^\pi_{T_l}(s,a)$ is large but $p^{\pi^\mu}_{T_t}(s,a)$ is small, for \textit{all} $s,a \in S,A$. As a result, the worst case bound becomes uncontrollably large when any rollouts from $T_l$ diverges too much from the behavior dataset.  
\\
\\
What happens then if our assumption on $C$ doesn't hold? This implies that the previous worst case error becomes unbounded. A simple algebraic modification in the proof, however resolves this issue, albeit at a cost of losing our convergence guarantees. But first we define a new term $U$, that attempts to capture an intuitive notion of uncertainty\cite{pessimportance} induced by the error in the learned model. We define $U(\pi) := 
|\EX_{s,a \sim p^{\pi}_{T_l}(s,a)}[||
T_t(s'|a,s) - T_l(s'|a,s)||_1] - \EX_{s,a \sim p^{\pi^{\mu}}_{T_t}(s,a)}[||
T_t(s'|a,s) - T_l(s'|a,s)||_1] |$, the difference in the model error induced by $\pi$ when evaluated at  $p^{\pi}_{T_l}(s,a)$ and $p^{\pi^{\mu}}_{T_t}(s,a)$. From its definition, we can observe that $\sup_{\pi \in \Pi}U(\pi)$ determines a property related to the global accuracy of $T_l$. We then utilize this term to bypass the unboundedness of $C$ in Theorem 2.
\\\\\textbf{Corollary 2:} \textit{Algorithm 2 returns a  policy $\hat{\pi}$ such that}
$$V^{\pi^*}  \leq V^{\hat{\pi}} +  \frac{1}{1-\gamma}2\epsilon_s + \frac{1}{(1-\gamma)^2}(2\epsilon_T + U(\hat{\pi}) + U(\pi^*)) + \frac{1}{1-\gamma}\epsilon_{\pi^*,\hat{\pi}} $$
\\
In words, the suboptimality increases when either $\hat{\pi}$ or $\pi^* $ visits states that incur  larger model error than $\epsilon_T$. Given that $T_l$ attempts to minimize $\epsilon_T$, we can expect that it is relatively more accurate on $s,a \sim p^{\pi^{\mu}}_{T_t}(s,a)$. Consequently, $U(\hat{\pi})$ and $U(\pi^*)$ is small when  $\hat{\pi}$ and $\pi^*$ doesn't erroneously lead to states that incur large model error. However, note that $U(\pi^*)$ is an uncontrollable term, and signifies a fundamental property of the provided dataset. \textit{Therefore any attempt to ensure the error is bounded should be realized by controlling $U(\hat{\pi})$. }
\\ \\ 
In practical settings, to avoid degenerate solutions caused by the potentially unbounded model bias, we instead want to solve the optimization problem of the following form, 
$$\underset{\pi \in \Pi}{\text{minimize}}\max_{f \in F} \frac{1}{M}\sum_{m=1}^{{M}} 
f(s^*) - \EX_{s \sim  d^{\pi}_{T_l}}[f(s)] +\frac{1}{M}\sum_{m=1}^{{M}}||\pi^*(a|s^*) - \pi(a|s^*)||_1  $$
$$ s.t.  U(\hat{\pi}) \leq c$$ 
where $c$ is some sufficiently small constant. 
An immediate question is then how best do we ensure a sufficiently small $U(\hat{\pi})$? By its definition, we know $U(\hat{\pi})$ will be small when the learner $\hat{\pi}$ visits states that the model is relatively more accurate in. Certainly, we cannot formulate this as a Lagrangian to directly minimize $U(\hat{\pi})$, as no such oracle exists. In section 5, we provide some practical remedies which proves to be essential to Algorithm 2's practical performance.

\section{Practical Implementation and Experiments}
Up to this point our theoretical results indicates that in the offline imitation learning setting, using a learned model from suboptimal datasets reduces the dependence on the number of expert samples, provided that the algorithm can control its degree of distributional shift(bounded $C$). Given that the results were based on certain idealizations and assumptions, can we still empirically validate this claim? 

\paragraph{Uncertainty Penalization}
In section 4 we discussed how the sample complexity of Algorithm 1 and 2 crucially depends on the uncertainty of $T_l$ in the states induced by $\hat{\pi}$ and $\pi^*$. Instead of relying on the assumption on $C$, we decompose the model's uncertainty into $U(\hat{\pi})$ and $U(\pi^*)$, and argue that while $U(\pi^*)$ is a fundamental property of the dataset, we can attempt to control the value of $U(\hat{\pi})$. In our experiments, we utilise uncertainty quantification techniques\cite{mopo, morel,combo,offlinetutorial} to estimate and control $U(\hat{\pi})$. Following works in the model based offline reinforcement learning literature\cite{mbpo}, we train $T_l$ as an ensemble of $K$ probabilistic neural networks that each outputs a mean and a covariance matrix $\Sigma$ to estimate the transition from the behavior dataset, and use the discrepancies in their predictions to estimate uncertainty. While there are a myriad of possible techniques for this endeavor, for our main evaluations we use the one presented in \cite{mopo} and observe that it is sufficient to obtain good results.  Specifically, we define the state-action wise estimated uncertainty $\hat{U}(s,a) := \max_{i=1}^{K}||\Sigma_{i}(s,a)||$ and the uncertainty penalized discriminator $f_u(s,a) = f(s,a) - \lambda\hat{U}(s,a)$
where the objective of the policy then becomes(with our previously defined $\hat{f}$)
$$\text{loss}(\hat{\pi}) = \frac{1}{M}\sum_{m=1}^{{M}}||\pi^*(a|s^*) - \hat{\pi}(a|s^*)||_1 - \EX_{s \sim  d^{\hat{\pi}}_{T_l}}[\hat{f_u}(s)]\ $$
\\ \\ 
In our experiments we attempt to answer the following questions: \textbf{i)} In the offline setting, does Algorithm 2 outperform BC in the low data regime? \textbf{ii)} Is Algorithm 2 competitive with Adversarial IL methods that are trained online? \textbf{iii)} To what extent does $U(\hat{\pi})$ affect the algorithm's performance and can we validate that we can indeed control $U(\hat{\pi})$ with the proposed methods? We answer the above questions with standard  benchmark tasks from OpenAi gym\cite{openai} with the Mujoco\cite{mujoco} simulator. We collect both expert and behavior datasets using SAC\cite{sac}. All behavior datasets are collected for 1M time steps, after which the collection terminates. We then evaluate our algorithm along with BC and Adversarial IL from section 3. We implement our own Adversarial IL algorithm and practice fair comparison by training it on an identical codebase and hyperparameters(where they apply) as on Algorithm 2, except that it uses a true model $T_t$ to sample rollouts instead. BC also shares the same policy architecture with Adversarial IL and Algorithm 2. 

We first highlight several factors that are expected to affect the magnitude of $U(\hat{\pi})$. They include: \\ 
\textbf{1) Dataset Size}: The number of state action pairs used to train the model\\
\textbf{2) Dataset Coverage}: The degree of exploration or the number of policies used to collect the data.\\
\textbf{3) Uncertainty Penalization}: This requires uncertainty quantification techniques.\\
\textbf{4) Starting State}: As alluded before, the starting state of rollouts directly affects the model's uncertainty.   \\ 
\\ Note that \textbf{1)} and \textbf{2)} are both 
determined prior running the algorithm. Assuming that \textbf{1)} and \textbf{2)} are known, we can use this prior to determine the degree of \textbf{3)}. In order to evaluate the severity of \textbf{2)}, we collect two different behavior datasets to train the model. Dataset $p^{\pi^\mu}_{w}$ is collected by a slow learning policy that upon termination, obtains medium level performance, or approximately half the returns of the $\pi^*$. We deliberately induce slow learning so that the best performing policy in $p^{\pi^\mu}_{w}$ is still suboptimal. Dataset $p^{\pi^\mu}_{n}$ is collected entirely by a random policy. Note that in the evaluated Mujoco Tasks, a random policy admits a much narrower dataset than a learning policy. Main experimental evaluations are shown in Table 1. Performance for Algororithm 2 was evaluated on the $T_l$ trained with the $p^{\pi^\mu}_{w}$\textbf{(Algorithm2_w on Table 1)} and  $p^{\pi^\mu}_{n}$\textbf{(Algorithm2_n on Table 1)} dataset. For all tasks, we experiment with varying number of expert trajectories, namely 1,3,10 and 30. When training the model, we utilise only the prescribed expert dataset(s) in addition to the behavior dataset.

We observe that Algorithm2_w \textit{always} outperforms BC with large margin on the 1,3,10 Expert Traj settings, while they both become near optimal when provided 30 expert trajectories. We notice that BC is more susceptible to $M$ in \textit{both} directions, and its performance admits a steeper slope as a function of $M$. This is confirmatory evidence to our analysis in that the dependency on $M$ is much lower for Algorithm 2. For $Algorithm2_n$, we obtain much lower performance than $Algorithm2_w$, which is expected due to lower coverage of $p^{\pi^\mu}_{n}$. However, it still outperforms BC on most settings. We also note that BC is significantly more susceptible to high variance among trials, which could be detrimental for real world settings\cite{bclimitation}. We also point out that our implementation of Adversarial IL aligns with the literature\cite{errorbound,imidiv,divergence,gail,airl,gcl} with respect to its performance, and also our analysis in that it is robust to low data regimes. Despite having to use a learned model instead, Algorithm 2 trained on $p^{\pi^\mu}_{w}$ is almost as good as Adversarial IL.
\\ \\ 
We also empirically study the effect of different starting states from section 5. Results are presented in Table 2. We find that the performance heavily relies on where $\hat{\pi}$ begin its rollouts on $T_l$. During experiments, we find that $T_l$'s low accuracy on states other than from $d^{\pi^\mu}_{T_t}(s)$ leads to unstable policy learning. This result aligns with our theoretical findings, and it suggests that different starting states could potentially lead to stronger model-based offline RL results. Additional implementation details/evaluations are provided in the Appendix.

\begin{table}
\centering
\label{t5}
\begin{tabular}{cc|cccc}
\noalign{\smallskip}\noalign{\smallskip}\hline\hline
\multicolumn{2}{c|}{} & BC & Adversarial IL & Algorithm2_w & Algorithm2_n\\
\hline
Hopper-v2 & 1 Expert Traj & 0.15 $\pm$ 0.16  & 0.83 $\pm$ 0.13  & 0.83 $\pm$ 0.11 & 0.52 $\pm$ 0.13 \\
& 3 Expert Traj & 0.14 $\pm$ 0.18 & 0.95 $\pm$ 0.03 & 0.78 $\pm$ 0.11 & 0.55 $\pm$ 0.21 \\
& 10 Expert Traj & 0.56 $\pm$ 0.25 & 1.02 $\pm$ 0.06 & 0.95 $\pm$ 0.13 & 0.61 $\pm$ 0.10 \\
& 30 Expert Traj & 0.91 $\pm$ 0.06 & 0.94 $\pm$ 0.09 & 0.95 $\pm$ 0.02 & 0.84 $\pm$ 0.04 \\
\hline
{HalfCheetah-v2} & 1 Expert Traj & 0.21 $\pm$ 0.33 & 0.72 $\pm$ 0.23 &  0.42 $\pm$ 0.08  & 0.33 $\pm$ 0.04 \\
& 3 Expert Traj & 0.55 $\pm$ 0.21 & 0.93 $\pm$ 0.04 & 0.72 $\pm$ 0.11 & 0.31 $\pm$ 0.12 \\
& 10 Expert Traj &  0.57 $\pm$ 0.12  & 1.13 $\pm$ 0.04&  0.78 $\pm$ 0.18& 0.52 $\pm$ 0.04 \\
& 30 Expert Traj & 0.83 $\pm$ 0.10 & 0.95 $\pm$ 0.02 & 0.93 $\pm$ 0.11 & 0.78 $\pm$ 0.11 \\
\hline
{Walker2d-v2} & 1 Expert Traj &  0.09 $\pm$ 0.12 &  0.74 $\pm$ 0.22 &  0.66 $\pm$ 0.20 & 0.32 $\pm$ 0.14 \\
& 3 Expert Traj & 0.10 $\pm$ 0.16 & 0.82 $\pm$ 0.09 & 0.73 $\pm$ 0.22 & 0.78 $\pm$ 0.11 \\
& 10 Expert Traj &  0.37 $\pm$ 0.42 &  0.91 $\pm$ 0.12 &  0.98 $\pm$ 0.03 & 0.89 $\pm$ 0.05 \\
& 30 Expert Traj & 0.92 $\pm$ 0.34 & 0.97 $\pm$ 0.13 & 0.98 $\pm$ 0.09 & 0.94 $\pm$ 0.05 \\
\hline
\hline
\end{tabular}
    \begin{tablenotes}
      \small
      \item Table 1: Mean and Std for performance of Algorithm 2 compared to BC and Adversarial IL evaluated on the last iteration, with varying number of provided expert trajectories. Scores are normalized such that a random policy scores 0.00 and the expert scores 1.00. All trials were averaged over different random seeds. For BC, the numbers presented are averaged over 20 different seeds due to a much shorter run time. Adversarial IL and Algorithm 2 were averaged over 3 seeds. For all 1 Expert Traj experiments, different trajectory was used each trial. 
    \end{tablenotes}
\end{table}

 \begin{table}
\label{t3}
\begin{tabular}{c|cccc}
\noalign{\smallskip}\noalign{\smallskip}\hline\hline
Starting States & $d_0(s)$ & $d^{\pi^*}_{T_t}(s)$ & $d^{\pi^\mu}_{T_t}(s)$ & $d(s)$ \\
\hline
Hopper-v2 & 0.21 $\pm$ 0.07 & 0.36 $\pm$ 0.05  & 0.78 $\pm$ 0.11 & 0.13 $\pm$ 0.05 \\
HalfCheetah-v2 & 0.32 $\pm$ 0.08 & 0.41 $\pm$ 0.05 &  0.72 $\pm$ 0.11 & 0.02 $\pm$ 0.01 \\
Walker2d-v2 & 0.18 $\pm$ 0.09 & 0.48 $\pm$ 0.21 & 0.73 $\pm$ 0.22 & 0.12 $\pm$ 0.04 \\
\hline
\hline
\end{tabular}
 \begin{tablenotes}
      \small
      \item Table 2: Algorithm2_w evaluated on different rollout starting states, all on the 3 Expert Traj settings. Scores are evaluated using the same procedure as in Table 1. 
    \end{tablenotes}
\end{table}

\section{Related Works}
We now briefly discuss prior works that motivated this work. Traditionally, IL focuses on learning with fixed expert demonstrations\cite{firstbc, bclimitation}, while \cite{dagger,reductions} formalized the compounding error bound of BC. Subsequently they introduce Dagger, which necessitates interactions with the expert and the environment, and shows that it breaks the quadratic bound. \cite{sail,sail2} present methods that address the distribution shift of IL by attempting to match the state visitation density of the expert. \cite{scalable,strictlybatch} recently proposed algorithms for offline imitation learning, but they are only restricted to discrete action spaces. 
\cite{gail,airl,gcl} first derived adversarial IL from the principle of maximum entropy \cite{maxent} in the deep RL setting, and demonstrated that it significantly outperforms BC. \\ \\
Here we highlight works that attempts IL with suboptimal demonstrations\cite{trex,suboptimal,failed,noisybc}. \cite{noisybc} presents an alternative BC algorithm that learns from noisy logged demonstrations, by filtering out data with low modes.\cite{trex} showed that, given performance rankings of the suboptimal demonstrations, their algorithm obtains better performance than the demonstration. However, their algorithm relies on the assumption that the reward in the MDP of interest is easily 'extrapolatable', which  is  not a fair assumption to make on all MDPs, such as ones that require a fair amount of exploration. \\\\
We also highlight some theoretical contributions that motivated our work, namely\cite{imilimits, fail, errorbound}.\cite{fail} presents a provably efficient imitation learning algorithm on settings where the actions are not provided in the expert dataset. In the tabular setting \cite{imilimits} showed that with knowledge of the true dynamics one can obtain an subquadratic error bound with respect to the horizon. \cite{errorbound} illustrates a unified view on error bound analysis for imitation learning. While \cite{errorbound,imilimits} does not present any practical algorithms, our theoretical results were greatly inspired by their works. In that sense, our work can be considered an extension of their works.
\\\\
 In its essence, our empirical work most resembles the recent works in model based offline learning\cite{combo,morel,mopo}, which have shown promising results. Therefore, any further developments in this area could potentially benefit our work. To combat the issue of distribution shift, \cite{mopo,morel} has relied on uncertainty penalization using the discrepancies in the ensemble predictions, while \cite{combo} has relied on penalizing the Q function for state action pairs that are generated by the model. In particular, \cite{mopo,combo} builds on the work of \cite{mbpo} that uses an ensemble of model to estimate epistemic uncertainty. Outside offline RL, \cite{kiante} also utilises ensemble disagreement of policies to penalize visiting uncertain regions.

\section{Conclusion}
Addressing sample efficiency and covariate shift has been a long standing challenge in IL. In our work, we present principled remedies by finding usage of suboptimal demonstrations. We show that training a model from a suboptimal dataset and adversarial training in the estimated MDP yields an algorithm that requires much lower amount of expert samples. As the algorithms we present are general and backed with theory, it could potentially have real world use cases. It shares the same promise with offline RL, but without requiring reward labels.\\ \\ 
It is also worthy to share some limitations in our work. Notably, our algorithm relies on the sufficient coverage assumption of the behavior dataset along with less principled approaches to estimate uncertainty. In that perspective, our work shares the same obstacles as model based offline RL algorithms, and future works on these limitations could benefit them both.

\section*{Checklist}

The checklist follows the references.  Please
read the checklist guidelines carefully for information on how to answer these
questions.  For each question, change the default \answerTODO{} to \answerYes{},
\answerNo{}, or \answerNA{}.  You are strongly encouraged to include a {\bf
justification to your answer}, either by referencing the appropriate section of
your paper or providing a brief inline description.  For example:
\begin{itemize}
  \item Did you include the license to the code and datasets? \answerYes{See Section~\ref{gen_inst}.}
  \item Did you include the license to the code and datasets? \answerNo{The code and the data are proprietary.}
  \item Did you include the license to the code and datasets? \answerNA{}
\end{itemize}
Please do not modify the questions and only use the provided macros for your
answers.  Note that the Checklist section does not count towards the page
limit.  In your paper, please delete this instructions block and only keep the
Checklist section heading above along with the questions/answers below.

\begin{enumerate}

\item For all authors...
\begin{enumerate}
  \item Do the main claims made in the abstract and introduction accurately reflect the paper's contributions and scope?
    \answerYes{}
  \item Did you describe the limitations of your work?
    \answerYes{see conclusion}
  \item Did you discuss any potential negative societal impacts of your work?
    \answerNA{no forseeable negative impact}
  \item Have you read the ethics review guidelines and ensured that your paper conforms to them?
    \answerYes{}
\end{enumerate}

\item If you are including theoretical results...
\begin{enumerate}
  \item Did you state the full set of assumptions of all theoretical results?
    \answerYes{See preliminaries}
	\item Did you include complete proofs of all theoretical results?
    \answerYes{in appendix}
\end{enumerate}

\item If you ran experiments...
\begin{enumerate}
  \item Did you include the code, data, and instructions needed to reproduce the main experimental results (either in the supplemental material or as a URL)?
    \answerNo{code will soon be released after cleanup}
  \item Did you specify all the training details (e.g., data splits, hyperparameters, how they were chosen)?
    \answerYes{in appendix}
	\item Did you report error bars (e.g., with respect to the random seed after running experiments multiple times)?
    \answerYes{see Table 1, Table 2 in section 5}
	\item Did you include the total amount of compute and the type of resources used (e.g., type of GPUs, internal cluster, or cloud provider)?
    \answerNo{all experiments were run on cpus, as it doesn't need heavy compute}
\end{enumerate}

\item If you are using existing assets (e.g., code, data, models) or curating/releasing new assets...
\begin{enumerate}
  \item If your work uses existing assets, did you cite the creators?
    \answerNA{}
  \item Did you mention the license of the assets?
    \answerNA{}
  \item Did you include any new assets either in the supplemental material or as a URL?
    \answerNA{}
  \item Did you discuss whether and how consent was obtained from people whose data you're using/curating?
    \answerNA{}
  \item Did you discuss whether the data you are using/curating contains personally identifiable information or offensive content?
    \answerNA{}
\end{enumerate}

\item If you used crowdsourcing or conducted research with human subjects...
\begin{enumerate}
  \item Did you include the full text of instructions given to participants and screenshots, if applicable?
    \answerNA{}
  \item Did you describe any potential participant risks, with links to Institutional Review Board (IRB) approvals, if applicable?
    \answerNA{}
  \item Did you include the estimated hourly wage paid to participants and the total amount spent on participant compensation?
    \answerNA{}
\end{enumerate}

\end{enumerate}

\appendix

\begin{algorithm}[H]
\SetAlgoLined
Input: Optimal trajectories  $D(\pi^*) = \{s^*_i, a^*_i\}_{i=1..M}$, Suboptimal trajectories $D(\tilde{\pi}) = \{\tilde{s}_i, \tilde{a}_i\}_{i=1..N}$\;
\KwResult{$\hat{\pi}$,$T_l$, and $f$ parameters}
Train model $T_l$ with $D := D(\pi^*) \cup 
D(\tilde{\pi})$\;
    \For{epoch =1,2,...,T}{
        
        \textbf{Collect Data}: Run current policy(in parallel) $\hat{\pi}$ in $T_l$ for $H$ horizon length and obtain dataset $D(\hat{\pi}) = \{\hat{s}_i,\hat{a}_i, \hat{r}_i\}_{i=1..B}$\ with batch size B and scores $\hat{r}$ from $f$;
        
        \textbf{Discriminator}: Update $f$ with $D(\hat{\pi})$ and $D(\pi^*)$; 
        
        \textbf{Policy}: Take policy gradient step to maximise  $\hat{r}$ from $f$ \textit{and} minimize BC loss from $D(\pi^*)$;

    }

\caption{Algorithm 2}
\end{algorithm}\medskip

\section{Additional Implementation Details}

\paragraph{Rollout Length}
For sake of simplicity our analysis relied on the assumption that $d^{\hat{\pi}}_{T_l}$ is computable for a given policy $\hat{\pi}$. However, with infinite state action spaces, naively rolling out the entire trajectory for $\frac{1}{1-\gamma}$ steps in $T_l$ might not be desirable\cite{mbpo} as the model's error compounds as shown in section 4\cite{rltheory}. Meanwhile too short of a rollout from $T_l$ such as in\cite{mopo,mbpo} increases the sampling bias from not sampling from $d^{\hat{\pi}}_{T_l}$, and the resulting algorithm strays too much from our analysis. In our experiments, we find that a certain range of rollout lengths $H$ works well, namely $5\leq H\leq 15$. However we perform all our experiments with $H$ fixed with 10, which is typically longer than most model based RL experiments. As opposed to standard model based RL\cite{mbpo}, in the imitation learning setting we provide some additional justifications as to why a longer $H$ (up to a certain point) might be desirable in section B.1.

\paragraph{Starting State}
As discussed before, the policy performs $H$ step rollouts in $T_l$ where we then estimate $d^{\hat{\pi}}_{T_l}$. However, from what state distribution do we actually begin the rollout? This stems a fairly important discussion, as our experiments showed that it is a large controlling factor on the algorithm's performance. We first discuss several available options:  \textbf{i)} $d_0(s)$; Initial States: we assume samples from $d_0(s)$ are available as initial states in the provided dataset, 
\textbf{ii)} $d^{\pi^*}_{T_t}(s)$; Expert States,
\textbf{iii)} $d^{\pi^\mu}_{T_t}(s)$; Behavior States,
\textbf{iiii)} $d(s)$; Arbitary States: these refers to any states that are sampled from the observation space. 

We first state that we chose \textbf{iii)} for our evaluations, motivated from our theoretical results. We also experimentally confirm that \textbf{iii)} is indeed the best option. In section 4, we showed that the algorithm's performance relies crucially on the magnitudes of the term $C$ or $U(\hat{\pi})$. While $C$ is a property of the dataset, $U(\hat{\pi})$ is a term that we can control. Now recall the definition of $U(\hat{\pi})$. Since we expect $T_l$ to be relatively more accurate on $d^{\pi^\mu}_{T_t}(s)$ than others, a policy that begins from  $d^{\pi^\mu}_{T_t}(s)$ will also likely end up in states that $T_l$ is more accurate in, which results in a lower $U(\hat{\pi})$. Explained differently, the algorithm's performance depends on the accuracy of $T_l$ during rollouts, which is expected to be higher when rollouts start from \textbf{iii)}. In a sense, choosing the starting state \textbf{iii)} is an additional strategy that controls $U(\hat{\pi})$, as is uncertainty penalization. We corroborate this hypothesis in our experiments in section 5. 
\\\\
\subsection{Additional Training Details}
\paragraph{Discriminator} We train the function $f$ with 2 layers and 64 units. To mitigate overfitting, we regularize it by \textit{enforcing a lipschitz constant} of 0.05 on the network parameters. We find this to be significantly more beneficial than gradient penalty based regularizations.
\paragraph{Model} We train the model as similar to \cite{mopo,mbpo}, where we train an ensemble of 7 neural networks and predict with a random sample from the ensemble predictions. 
\paragraph{Policy} We optimize the policy with PPO\cite{ppo}. Although we can in principle backpropagate through time in $T_l$, we believed policy gradients would be a better choice since we can get a gradient arbitarily close to the true gradient by simply sacrificing more compute. For all our experiments, we collect 5000 samples in parallel for a horizon length of 10, which results in 50000 total samples per epoch. 

\section{Proofs}

\textbf{Proof of Lemma 1}(Performance Difference of Two Policies):
This is a simple application of the performance difference lemma\cite{rltheory} of any two policies $\pi$, $\hat{\pi}$. Let $R_{max} \leq 1/2$,
then it follows that  $Q_{max} \leq \frac{1}{2(1-\gamma)}$, where $Q^\pi(s,a) := \EX_{ \pi, T_t}[\sum_{t=0}^{\infty}\gamma^t r(s,a)] $ and $Q^\pi(s,\hat{\pi}) := \EX_{a\sim \hat{\pi}, \pi,T_t}[\sum_{t=0}^{\infty}\gamma^t r(s,a)] $, where only the first action is averaged over $\hat{\pi}$. Then it follows that 

$$V^\pi - V^{\hat{\pi}}$$ 
$$= \frac{1}{1-\gamma}\EX_{s\sim d^\pi_{T_t}}[-Q^{\hat{\pi}}(s,\hat{\pi}(s)) + Q^{\hat{\pi}}(s,\pi(s))]$$

$$\leq Q_{max}\frac{1}{1-\gamma}\EX_{s\sim d^\pi_{T_t}}[||\pi - \hat{\pi}||_1] $$
$$\leq \frac{1}{(1-\gamma)^2} \EX_{s\sim d^\pi_{T_t}}[||\pi - \hat{\pi}||_1] $$ 

We refer readers to Section E, Theorem 21 in \cite{agarwal} for detailed proofs on maximum likelihood(MLE) guarantees on $\EX_{s\sim d^\pi_{T_t}}[||\pi - \hat{\pi}||_1]$, such that the bound becomes a valid sample complexity after a MLE training procedure. 
\\ \\ 
\textbf{Proof of Theorem 2}: The proof can be found in \cite{rltheory}, but we also present here for completeness. It is a direct application of a uniform convergence argument for all $f \in F$. For Theorem 2 and 3, we define $f$ slightly differently from its definition presented in section 2, namely $f$ takes both $s,a$ as inputs. We then define $\hat{f} := \argmax_{f \in F} \frac{1}{M}\sum_{m=1}^{{M}} 
f(s^*,a^*) - \EX_{s \sim  p^{\hat{\pi}}_{T_t}}[f(s,a)]$ and $\tilde{f} := \argmax_{f \in F} \EX_{s \sim  p^{\pi^*}_{T_t}}[f(s,a)] - \EX_{s \sim  p^{\hat{\pi}}_{T_t}}[f(s,a)]$. 

Then it follows that for all $\pi \in \Pi$, $\max_{f \in F}(\EX_{s,a \sim  p^{\pi^*}_{T_t}}[f(s,a)] - \EX_{s,a\sim  p^{\hat{\pi}}_{T_t}}[f(s,a)]) = \max_{f: ||f||_{\infty}\leq 1}(\EX_{s,a \sim  p^{\pi^*}_{T_t}}[f(s,a)] - \EX_{s \sim  p^{\hat{\pi}}_{T_t}}[f(s,a)]) = ||p^{\pi^*}_{T_t}(s,a) - p^{\hat{\pi}}_{T_t}(s,a)||_1$ \\ \\ 
Also, let $R_{max} \leq 1$. It is then straightforward to see that 
$$ V^{\pi^*} - V^{\hat{\pi}} $$
$$\leq \frac{1}{1-\gamma}|||p^{\pi^*}_{T_t}(s,a) - p^{\hat{\pi}}_{T_t}(s,a)||_1 $$
\\
We now bound the distance of the state-action visitation densities, 
$$||p^{\pi^*}_{T_t}(s,a) - p^{\hat{\pi}}_{T_t}(s,a)||_1$$ 
$$= \EX_{s,a \sim  p^{\pi^*}_{T_t}}[\tilde{f}(s,a)] - \EX_{s \sim  p^{\hat{\pi}}_{T_t}}[\tilde{f}(s,a)]$$
$$\leq \frac{1}{M}\sum_{m=1}^{{M}} 
\tilde{f}(s^*,a^*) - \EX_{s,a \sim  p^{\hat{\pi}}_{T_t}}[\tilde{f}(s,a)] + \epsilon_s$$
$$\leq \frac{1}{M}\sum_{m=1}^{{M}} 
\hat{f}(s^*,a^*) - \EX_{s,a \sim  p^{\hat{\pi}}_{T_t}}[\hat{f}(s,a)] + \epsilon_s$$
$$\leq \frac{1}{M}\sum_{m=1}^{{M}} 
\hat{f}(s^*,a^*) - \EX_{s,a \sim  p^{\pi^*}_{T_t}}[\hat{f}(s,a)] + \epsilon_s$$
$$\leq 2\epsilon_s$$
Utilising
$$ V^{\pi^*} - V^{\hat{\pi}} $$
$$\leq \frac{1}{1-\gamma}||p^{\pi^*}_{T_t}(s,a) - p^{\hat{\pi}}_{T_t}(s,a)||_1 $$ we conclude our proof.
 The first and last inequality is from Hoeffding's inequality taken union bound for all $f \in F$, where with probability at least $1-\delta$ , $ | \EX_{s \sim p^{\pi^*}_{T_t}(s,a)}[f(s,a)]
- \frac{1}{M}\sum_{m=1}^{{M}} f(s^*,a^*)| \leq 
2\sqrt{\frac{ln(\frac{|F|}{\delta})}{M}},  \forall  f \in F$, $\epsilon_s := 2\sqrt{\frac{ln(\frac{|F|}{\delta})}{M}}$, and the third inequality is from the optimality of $\hat{\pi}$.
\\ \\ 
\textbf{Proof of Theorem 3}:
The proof is an adaptation of Theorem 2 in the case where the expectation over $ p^{\pi^*}_{T_t}$ is replaced with $ p^{\pi^*}_{T_l}$, then with the same definitions as in the proof of Theorem 2,  
$$||p^{\pi^*}_{T_t}(s,a) - p^{\hat{\pi}}_{T_t}(s,a)||_1$$ 
$$= \EX_{s,a \sim  p^{\pi^*}_{T_t}}[\tilde{f}(s,a)] - \EX_{s,a \sim  p^{\hat{\pi}}_{T_t}}[\tilde{f}(s,a)]$$
$$\leq \frac{1}{M}\sum_{m=1}^{{M}} 
\tilde{f}(s^*,a^*) - \EX_{s,a \sim  p^{\hat{\pi}}_{T_t}}[\tilde{f}(s,a)] + \epsilon_s $$
$$\leq \frac{1}{M}\sum_{m=1}^{{M}} 
\tilde{f}(s^*,a^*) - \EX_{s,a \sim  p^{\hat{\pi}}_{T_l}}[\tilde{f}(s,a)] + \epsilon_s + \frac{C}{1-\gamma}\epsilon_T$$
$$\leq \frac{1}{M}\sum_{m=1}^{{M}} 
\tilde{f}(s^*,a^*) - \EX_{s,a \sim  p^{\pi^*}_{T_l}}[\tilde{f}(s,a)] + \epsilon_s + \frac{C}{1-\gamma}\epsilon_T$$
$$\leq \frac{1}{M}\sum_{m=1}^{{M}} 
\tilde{f}(s^*,a^*) - \EX_{s,a \sim  p^{\pi^*}_{T_t}}[\tilde{f}(s,a)] + \epsilon_s + \frac{2C}{1-\gamma}\epsilon_T$$
$$\leq \frac{1}{M}\sum_{m=1}^{{M}} 
\tilde{f}(s^*,a^*) - \EX_{s,a \sim  p^{\pi^*}_{T_t}}[\tilde{f}(s,a)] + \epsilon_s$$
$$\leq 2\epsilon_s + \frac{2C}{1-\gamma}\epsilon_T$$

The third inequality is from the optimality of $\hat{\pi}$. For the second and fourth inequality  we use lemma 6 presented below to bound  $||d^{\hat{\pi}}_{T_t}(s) - d^{\hat{\pi}}_{T_l}(s)||_1 \leq \frac{1}{1-\gamma}\epsilon_T$. Consequently, using
$$ V^{\pi^*} - V^{\hat{\pi}} $$
$$\leq \frac{1}{1-\gamma}||p^{\pi^*}_{T_t}(s,a) - p^{\hat{\pi}}_{T_t}(s,a)||_1 $$ we conclude our proof

\textbf{Proof of Theorem 4}:

We can decompose the state action distribution distance by the following, 
$$||p^{\pi^*}_{T_t}(s,a) -p^{\hat{\pi}}_{T_t}(s,a) ||_1 $$ 
$$\leq ||d^{\pi^*}_{T_t}(s) - d^{\hat{\pi}}_{T_t}(s)||_1 + 
\EX_{s \sim d^{\pi^*}_{T_t}(s)}[||\pi^*(a|s) - \hat{\pi}(a|s)||_1]
$$
\\
Define $$\hat{f} = \argmax_{f \in F}\EX_{s \sim d^{\pi^*}_{T_t}(s)}[f(s)] - \EX_{s \sim d^{\hat{\pi}}_{T_t}(s)}[f(s)]$$
 $$ \tilde{f} = \argmax_{f \in F} \frac{1}{M}\sum_{m=1}^{{M}} f(s^*) - \EX_{s \sim d^{\hat{\pi}}_{T_l}(s)}[f(s)]  $$ \\ Using Hoeffding's inequality and taking the union bound for all discriminator functions $f \in F$(where we have defined in section 2), we have that with probability at least $1-\delta$,
$$ | \EX_{s \sim d^{\pi^*}_{T_t}(s)}[f(s)]
- \frac{1}{M}\sum_{m=1}^{{M}} f(s^*)| \leq 
2\sqrt{\frac{ln(\frac{|F|}{\delta})}{M}},  \forall  f \in F$$\\
Now, bounding the difference in the two state distributions, 
$$||d^{\pi^*}_{T_t} - d^{\hat{\pi}}_{T_t}||_1 $$
$$= \EX_{s \sim d^{\pi^*}_{T_t}(s)}[\hat{f}(s)] - \EX_{s \sim d^{\hat{\pi}}_{T_t}(s)}[\hat{f}(s)]$$
$$\leq \frac{1}{M}\sum_{m=1}^{{M}} \hat{f}(s^*) - 
\EX_{s \sim d^{\hat{\pi}}_{T_t}(s)}[\hat{f}(s)] + \epsilon_{s}$$
$$\leq \frac{1}{M}\sum_{m=1}^{{M}} \hat{f}(s^*) - 
\EX_{s \sim d^{\hat{\pi}}_{T_l}(s)}[\hat{f}(s)] + \epsilon_{s} + \frac{C\epsilon_T}{1-\gamma}$$
$$= \frac{1}{M}\sum_{m=1}^{{M}} \hat{f}(s^*) - 
\EX_{s \sim d^{\hat{\pi}}_{T_l}(s)}[\hat{f}(s)] + \epsilon_{s} + \frac{C\epsilon_T}{1-\gamma} + \frac{1}{M}\sum_{m=1}^{{M}}||\pi^*(a|s^*) - \hat{\pi}(a|s^*)||_1- \frac{1}{M}\sum_{m=1}^{{M}}||\pi^*(a|s^*) - \hat{\pi}(a|s^*)||_1$$
$$\leq \frac{1}{M}\sum_{m=1}^{{M}} \tilde{f}(s^*) - 
\EX_{s \sim d^{\hat{\pi}}_{T_l}(s)}[\tilde{f}(s)] + \epsilon_{s} + \frac{C\epsilon_T}{1-\gamma} + \frac{1}{M}\sum_{m=1}^{{M}}||\pi^*(a|s^*) - \hat{\pi}(a|s^*)||_1 - \frac{1}{M}\sum_{m=1}^{{M}}||\pi^*(a|s^*) - \hat{\pi}(a|s^*)||_1$$
$$\leq \frac{1}{M}\sum_{m=1}^{{M}} \tilde{f}(s^*) -
\EX_{s \sim d^{\pi^*}_{T_l}(s)}[\tilde{f}(s)] + \epsilon_{s} + \frac{C\epsilon_T}{1-\gamma}  - \frac{1}{M}\sum_{m=1}^{{M}}||\pi^*(a|s^*) - \hat{\pi}(a|s^*)||_1 $$
$$\leq \frac{1}{M}\sum_{m=1}^{{M}} \tilde{f}(s^*) -
\EX_{s \sim d^{\pi^*}_{T_t}(s)}[\tilde{f}(s)] + \epsilon_{s} + \frac{2C\epsilon_T}{1-\gamma} - \frac{1}{M}\sum_{m=1}^{{M}}||\pi^*(a|s^*) - \hat{\pi}(a|s^*)||_1 $$
$$\leq 2\epsilon_{s} + \frac{2C\epsilon_T}{1-\gamma} - \frac{1}{M}\sum_{m=1}^{{M}}||\pi^*(a|s^*) - \hat{\pi}(a|s^*)||_1 $$
Where the first equality holds from our definition of the set of discriminators $F$. For the second and sixth inequality we again use lemma 6. The fifth inequality holds due to the optimality of $\hat{\pi}$. 
Returning to our original bound on expected returns, $$ V^{\pi^*} - V^{\hat{\pi}} $$
$$\leq \frac{1}{1-\gamma}||p^{\pi^*}_{T_t}(s,a) - p^{\hat{\pi}}_{T_t}(s,a)||_1 $$
$$\leq \frac{1}{1-\gamma} (||d^{\pi^*}_{T_t}(s) - d^{\hat{\pi}}_{T_t}(s)||_1 + 
\EX_{s \sim d^{\pi^*}_{T_t}(s)}[||\pi^*(a|s) - \hat{\pi}(a|s)||_1])
$$
$$\leq \frac{1}{1-\gamma} (2\epsilon_{s} + \frac{2C\epsilon_T}{1-\gamma}  + \epsilon_\pi )$$

where we have $|\EX_{s \sim d^{\pi^*}_{T_t}(s)}[||\pi^*(a|s) - \hat{\pi}(a|s)||_1] - \frac{1}{M}\sum_{m=1}^{{M}}||\pi^*(a|s^*) - \hat{\pi}(a|s^*)||_1| \leq 2\sqrt{\frac{ln(\frac{1}{\delta})}{M}} := \epsilon_\pi $ with probability at least $1-\delta$.  \\ 
\\ \\ \textbf{Proof of Corollary 2} 
This follows from a slight modification in the proof of Theorem 4, where we bound $$|\EX_{s \sim d^{\hat{\pi}}_{T_l}(s)}[\hat{f}(s)] - \EX_{s \sim d^{\hat{\pi}}_{T_t}(s)}[\hat{f}(s)]| $$
$$\leq \frac{1}{1-\gamma} \EX_{d^{\hat{\pi}}_{T_l}}[||T_t(s'|s,a) - T_l(s'|s,a) ||_1 ]\leq \frac{1}{1-\gamma}
(\epsilon_T + U(\hat{\pi}))$$ and equivalently for $\pi^*$
\\ 
 \\ 
\textbf{Lemma 6 (Error Propagations of difference in transitions and policies)}: \cite{errorbound} has presented the state distribution error propagation bound for two differing policies. The bound for two differing transitions is trivially similar, and we present it here. We denote the transition operator $P^\pi_{T} := \sum_{a}T(s'|s,a)\pi(a|s)$ and $G^\pi_{T} := (\mathcal{I} - \gamma P^\pi_{T})^{-1}$, where we then have $(1-\gamma)G^\pi_{T_t}d_0  =d^{\pi}_{T_t}  $
\\ \\ 
Then it follows that, for two transitions $T_l$ and $T_t$, 
$$d^{\pi}_{T_t}(s) - d^{\pi}_{T_l}(s) $$
$$(1-\gamma)((\mathcal{I} - \gamma P^\pi_{T_t})^{-1} - (\mathcal{I} - \gamma P^\pi_{T_l})^{-1})d_0$$
$$ = (1-\gamma)\gamma G^\pi_{T_l}(P^\pi_{T_l} - P^\pi_{T_t})G^\pi_{T_t} d_0 $$
$$ =\gamma G^\pi_{T_l}(P^\pi_{T_l} - P^\pi_{T_t})d^{\pi}_{T_t} $$ \\ 
Where we then have  
$$ || d^{\pi}_{T_t}(s) - d^{\pi}_{T_l}(s) ||_1 =|| \gamma G^\pi_{T_l}(P^\pi_{T_l} - P^\pi_{T_t})d^{\pi}_{T_t}||_1 $$
$$\leq \gamma ||G^\pi_{T_t} ||_1 ||(P^\pi_{T_l} - P^\pi_{T_t})d^{\pi}_{T_t} ||_1  $$
where we can further bound the two terms individually, 
$$||G^\pi_{T_t} ||_1 \leq \frac{1}{1-\gamma} $$ and 
$$ ||(P^\pi_{T_l} - P^\pi_{T_t})d^{\pi}_{T_t}||_1 $$
$$ = \sum_{s^'}\sum_{s}(|P^\pi_{T_l}(s'|s) - P^\pi_{T_t}(s'|s)|)d^{\pi}_{T_t}(s)$$
$$= \sum_{s^'}\sum_{s}d^{\pi}_{T_t}(s) |\sum_{a}(T_t(s'|s,a) - T_l(s'|s,a))\pi(a|s) | $$
$$\leq \sum_{s^'}\sum_{s}\sum_{a}p^{\pi}_{T_t}(s,a)|T_t(s'|s,a) - T_l(s'|s,a) |  $$
$$= \EX_{p^{\pi}_{T_t}}[||T_t(s'|s,a) - T_l(s'|s,a) ||_1 ] $$
Then it follows that 
$$|| d^{\pi}_{T_t}(s) - d^{\pi}_{T_l}(s) ||_1 \leq \frac{1}{1-\gamma}\EX_{p^{\pi}_{T_t}}[||T_t(s'|s,a) - T_l(s'|s,a) ||_1 ]$$
\\
\paragraph{State Visitation difference minimization} We previously mentioned that $\hat{\pi}$ is minimizing the approximated state visitation difference, we clarify this notion here. 
$$||d^{\pi^*}_{T_t}(s) - d^{\hat{\pi}}_{T_t}(s)||_1 $$ 
$$
\leq ||d^{\pi^*}_{T_t}(s) - d^{\hat{\pi}}_{T_l}(s)||_1  + ||d^{\hat{\pi}}_{T_t}(s) - d^{\hat{\pi}}_{T_l}(s)||_1 
$$
$$
\leq ||d^{\pi^*}_{T_t}(s) - d^{\hat{\pi}}_{T_l}(s)||_1 +  \frac{1}{1-\gamma}C\epsilon_T $$ 
$$
 \leq \frac{1}{M}\sum_{m=1}^{{M}} 
\hat{f}(s^*) - \EX_{s \sim  d^{\hat{\pi}}_{T_l}}[\hat{f}(s)]+\frac{1}{1-\gamma}C\epsilon_T + \epsilon_s$$
The last inequality follows from our definition of $\hat{f}$ in Theorem 4. Intuitively, the distributional information of states is encoded in $\hat{f}$, which $\hat{\pi}$ can exploit by maximising $\hat{f}$. Maximising $\hat{f}$ becomes identical to minimizing $||d^{\pi^*}_{T_t}(s) - d^{\hat{\pi}}_{T_t}(s)||_1$ when the model training error $\epsilon_T$ is 0, or equivalently when there is an infinite number of behavior state action pairs. 
\\ \\  \\ 
\subsection{Recovery}

In this subsection we comment on a direct consequence of our choice in the starting state, namely the recovery behavior of our algorithm, which solidifies our intuition on its empirical success over behavioral cloning. Imitation Learning also concerns the problem of mitigating covariate shift, which occurs when the learner deviates from the expert's state distribution $d^{\pi^*}_{T_t}$ during the evaluation stage. Naive BC methods does not prepare the policy to recover to $d^{\pi^*}_{T_t}$ in such instances. We argue that our model based approach acts to resolve this issue. 
\\
\\ 
We define $v_{\pi}^k$ as an unique recovery variable for every policy $\pi$ in an MDP of interest. $$v_{\pi}^k := ||d^{\pi^*}_{T_t} - (P^{\pi}_{T_t})^k \nu ||$$ where $P^{\pi}_{T}$ is the transposed transition matrix $\sum_{a \in A}T(s'|a,s)\pi(a|s)$ induced by  $\pi$ in the dynamics $T$, and $\nu(s)$ denotes the density of any admissible states by $\pi \in \Pi$. Under the assumption of irreducibility and aperiodicity of  of $P^{\pi}_{T_t}$, $(P^{\pi}_{T_t})^k \nu(s)$ will converge to its stationary distribution $d^{\pi}_{T_t}(s)$ as $k \rightarrow{\infty}$.  Intuitively, the finiteness of $k$ allows $v_{\pi}^k$ to represent the ability of $\pi$ to 'direct' itself to $d^{\pi^*}_{T_t}(s)$ in under  $k$ transitions, starting from any admissible state $s \sim \nu(s)$. A policy that is sufficiently trained to minimize this objective could endow $\pi$ the ability to recover from any arbitrary admissible state back to the expert distribution in finitely many steps. It then follows that we can bound $v_{\pi}^k$ with certain terms of interest. 
\\
\\
\textbf{Proposition 1}
$$v_{\pi}^k \leq ||d^{\pi^*}_{T_t}(s) - 
(P^{\pi}_{T_l})^k d^{\pi^\mu}(s) || + k\hat{\epsilon_T} + \sqrt{2\log C}$$ 
Where $\hat{\epsilon_T}=\max_{i=1..k}\EX_{s,a \sim p^{\pi^\mu}(s,a)}[||
T_t(s_{i+1}|a_{i},s_{i}) - T_l(s_{i+1}|a_{i},s_{i})||_1]$. The first two terms of this bound are approximately the minimization variables in Algorithm 2. While we do not directly minimize $C$, we attempt to control a related variable $U(\hat{\pi})$ to ensure $C$ does not get too large. 
\\ \\ 
\textbf{Proof of Proposition 1}: 
$$||d^{\pi^*}_{T_t} - (P^{\pi}_{T_t})^k \nu ||_1$$
$$ = ||d^{\pi^*}_{T_t} - (P^{\pi}_{T_l})^k d^{\pi^\mu}+ (P^{\pi}_{T_l})^k d^{\pi^\mu} - (P^{\pi}_{T_t})^k d^{\pi^\mu} + (P^{\pi}_{T_t})^k d^{\pi^\mu} -  (P^{\pi}_{T_t})^k \nu ||_1$$
$$ \leq ||d^{\pi^*}_{T_t} - (P^{\pi}_{T_l})^k d^{\pi^\mu}||_1 + ||(P^{\pi}_{T_l})^k d^{\pi^\mu} - (P^{\pi}_{T_t})^k d^{\pi^\mu} ||_1 + || (P^{\pi}_{T_t})^k d^{\pi^\mu} -  (P^{\pi}_{T_t})^k \nu ||_1 $$
$$\leq  ||d^{\pi^*}_{T_t} - (P^{\pi}_{T_l})^k d^{\pi^\mu}||_1  + k\hat{\epsilon_T} + ||(P^{\pi}_{T_t})^k||_1||d^{\pi^\mu} - \nu||_1 $$
$$\leq ||d^{\pi^*}_{T_t} - (P^{\pi}_{T_l})^k d^{\pi^\mu}||_1  + k\hat{\epsilon_T} + \sqrt{2\log C}
$$

In the second inequality, we bound the $k$ step induced state divergence  from two different transition $T_l$ and $T_t$ $||(P^{\pi}_{T_l})^k d^{\pi^\mu} - (P^{\pi}_{T_t})^k d^{\pi^\mu} ||$ using the second inequality follows from applying lemma B.2 from \cite{mbpo}. Also note that $||(P^{\pi}_{T_t})^k||_1 = 1$ as $P^{\pi}_{T_t}$ is a left-stochastic matrix. The last inequality follows from Pinsker's inequality and the definition of $C$. We also point out a trade-off  between $k$ and $\hat{\pi}$'s recovery property. As $k$ decreases it is quite possible that $\hat{\pi}$ fails to learn to direct itself to $d^{\pi^*}_{T_t}$ from \textit{any} $s \sim \nu(s)$, especially if the  MDP of interest has a large state space. However, increasing $k$ loosens the bound for any non-zero $\epsilon_T$. This may explain the range of horizon $H$ that we found works well, as explained in section A.\\\\
Recall from section A that $\hat{\pi}$ begin its rollouts from $d^{\pi^\mu}_{T_t}(s)$. We then provide a consequential argument that arises from proposition 1; if $d^{\pi^\mu}_{T_t}(s)$ sufficiently represents the state space that is generated by a learner $\hat{\pi}$ from making a mistake, or equivalently when $C$ is bounded, then given a sufficiently accurate $T_l$, \textit{our practical instatiation of Algorithm 2 trains $\hat{\pi}$ to direct itself back to $d^{\pi^*}_{T_t}(s)$ in $k$ transitions.}
\\\\

\subsection{Reduction to offline RL}

In this section we study the setting where we are given additional information, namely \textbf{i)} when the true reward function is given and \textbf{ii}) when its empirical samples are given corresponding to the behavior dataset. Given the true reward function or labels, the problem then reduces to offline RL. In \textbf{i}, the policy $\pi$ generates rollouts in the learned dynamics $T_l$ with the true reward function $r: S \times A \mapsto \mathbb{R}$, after which $\pi$ maximises its collected reward in this fake environment. In \textbf{ii}, the setting is identical to \textbf{i} except that its reward function $\hat{r}: S \times A \mapsto \mathbb{R} $ is an empirical risk minimizing function of reward labels obtained from the behavior dataset $\{s_i, a_i, r_{s_i,a_i}\}_{i=1..N} \sim p^{\pi^\mu}_{T_t}$. This is in contrast to the imitation learning setting where we don't have access to either $r_{s_i, a_i}$ or $r$. Despite the simple analysis, it is insightful to compare how the bound compares to the previous setting under the same assumptions, since we can observe what differs when the true reward is replaced with a  discriminator function $f$. 

\textbf{Theorem 5(When True Reward Function is Given(i))}: Let 
$$ \hat{\pi} = \argmax_{\pi \in \Pi}\EX_{s,a \sim p^{\pi}_{T_l}(s,a)}[r(s,a)]$$
Then $$V^{\pi^*} \leq V^{\hat{\pi}} + \frac{2C\epsilon_T}{(1-\gamma)^2}$$ 

An almost identical proof is used for \textbf{ii}, where we take into account the approximation error for the reward function, $\epsilon_r :=
\EX_{s,a \sim p^{\pi^\mu}_{T_t}}[|\hat{r}(s,a) - r(s,a)|]$, so we just state the proof for \textbf{ii}. 
\\ 
\\
\textbf{Corollary 4(When Only Reward Labels are Given(ii))}: Let 
$$ \hat{\pi} = \argmax_{\pi \in \Pi}\EX_{s,a \sim p^{\pi}_{T_l}(s,a)}[\hat{r}(s,a)]$$
Then $$V^{\pi^*} \leq V^{\hat{\pi}} + \frac{2C\epsilon_T}{(1-\gamma)^2} + \frac{2C\epsilon_r}{1-\gamma}$$
\textbf{\\ Proof of Theorem 5/Corollary 4}
$$(1-\gamma)(V^{\pi^*} - V^{\hat{\pi}}) = 
\EX_{p^{\pi^*}_{T_t}(s,a)}[r(s,a)] - \EX_{p^{\hat{\pi}}_{T_t}(s,a)}[r(s,a)]$$ 
$$ \leq \EX_{p^{\pi^*}_{T_t}(s,a)}[r(s,a)] - \EX_{p^{\hat{\pi}}_{T_l}(s,a)}[\hat{r}(s,a)] + \frac{C\epsilon_T}{1-\gamma}  + C\epsilon_r
$$$$ \leq  \EX_{p^{\pi^*}_{T_t}(s,a)}[r(s,a)] - \EX_{p^{\pi^*}_{T_l}(s,a)}[\hat{r}(s,a)] + \frac{C\epsilon_T}{1-\gamma}  + C\epsilon_r
$$
$$ 
 \leq  \EX_{p^{\pi^*}_{T_t}(s,a)}[r(s,a)] - \EX_{p^{\pi^*}_{T_t}(s,a)}[r(s,a)] + \frac{2C\epsilon_T}{1-\gamma}  + 2C\epsilon_r = 
 \frac{2C\epsilon_T}{1-\gamma}  + 2C\epsilon_r 
$$

Compared to Theorem 3, given a true reward function, we are able to purge the high probability error term $\epsilon_s$. As in the imitation learning setting, Note that for both \textbf{i} and \textbf{ii} the suboptimality is still dependent on the concentrability coefficient. Analogous to Theorem 3 and 4, for \textbf{ii} the dependence on $\epsilon_T$ significantly outweighs that of the reward function approximation error $\epsilon_r$, which signifies the importance of an accurate model and data coverage over the reward function.

 \section{Additional Experiments}

In section 5 we saw that Algorithm 2 outperforms BC on all domains. In this section we provide some additional empirical analysis that serves to answer the following questions; 

 i) Compared to BC, does Algorithm 2 mitigate compounding errors?
\\ 
ii) Compared to BC, does the policy trained with Algorithm 2 admit the recovery behavior outlined in section B.1? \\ 
iii) How does Algorithm 2 perform when the model is trained \textit{solely} from expert demonstrations? \\  

In order to evaluate i), we use a separate model \textit{$\hat{T_l}$ that is trained solely from expert demonstrations} $d^{\pi^*}_{T_t}(s)$, and estimate its uncertainty when both $\hat{\pi}$ and $ \pi^{BC}$ perform rollouts in $T_l$. $\hat{\pi}$ and $\pi^{BC}$ is the output policy of Algorithm 2 and behavioral cloning, respectively.  Both $\hat{\pi}$ and $\pi^{BC}$ begin their rollouts in $T_l$ from the same sampled state $s^* \sim d^{\pi^*}_{T_t}(s)$(where $\hat{T_l}$ admits low uncertainty), and by quantifying its rise throughout the rollout, we estimate which policy becomes more divergent from $d^{\pi^*}_{T_t}(s)$. (recall $\hat{T_l}$ only encountered $d^{\pi^*}_{T_t}(s)$ while training). In particular, we estimate both $\hat{T_l}$'s  aleatoric and epistemic uncertainty by measuring  $$\tilde{U}_{t}(s_t,a_t;T_l) := \alpha\max_{i=1..k}||\tilde{\Sigma}_{i}(s,a)|| - \beta\log P(s'_{pred}|s_t,a_t)$$
$\log P(s'_{pred}|s,a)$ is $\hat{T_l}$'s estimated log-probability of the next state generated by $T_l$, $\tilde{\Sigma}_{i}(s,a)$ is the covariance matrix predicted by each member of the ensemble $\hat{T_l}$, and $\alpha$ and $\beta$ are appropriate scaling terms. $t$ indexes the timesteps elapsed after beginning the rollouts. 
\\ \\ 
Similarly, we evaluate ii) by letting both both $\hat{\pi}$ and $\pi^{BC}$ begin their rollouts from the same sampled state from $d^{\pi^\mu}_{T_t}(s)$, i.e $\tilde{s} \sim d^{\pi^\mu}_{T_t}(s)$. Indeed, $\hat{T_l}$ was not trained on $d^{\pi^\mu}_{T_t}(s)$ so it will admit high $\tilde{U}_{t}$ on $\tilde{s}$ at $t=1$. However we predict that a policy that admits  recovery behavior will learn to direct itself back towards $d^{\pi^*}_{T_t}(s)$ regardless of the starting state, which would in turn correlate with a decreasing $\tilde{U_t}$ along $t$. 
\\ \\
We provide confirmatory evidence of this hypothesis in figure 1 where we've outlined results for evaluating i) and ii). Recall that $\hat{\pi}$ and $\pi^{BC}$ generates rollouts in $T_l$. For each chart the x-axis represents the elapsed timestep $t$ during the rollout after initial sampling, and the y-axis represents the uncertainty $\tilde{U}_t$ as measured by $\hat{T_l}$. The lines for $\hat{\pi}$ and $\pi^{BC}$ are denoted by Algorithm2 and BC respectively. \textit{As the lines deviate more from the x-axis, the evaluated policy is visiting more uncertain states, which implies increased divergence from $d^{\pi^*}_{T_t}(s)$}, and vice versa. \\ \\ 
For i), the results are presented in the first two charts. Note that the BC agent admits a much more divergent behavior than Algorithm 2 when they both begin $s \sim d^{\pi^*}_{T_t}(s)$. The third and fourth charts outline the evaluation results for ii). As opposed to i), both BC and Algorithm 2 begins its rollouts from a state sampled from $d^{\pi^\mu}_{T_t}(s)$. As a result, the initial uncertainty measured by $T_l$ is greater than in i). We observe that in ii) $\pi^{BC}$ diverges even more so than in i) but $\hat{\pi}$ succeeds in visiting states that obtain gradually lower uncertainty. This result provides evidence to $\hat{\pi}$'s recovery behavior. We also remind the readers that as $T_l$ becomes more accurate, this comparison would correlate greater with the true divergence behavior in $T_t$.

\paragraph{Expert Only} So far, we've only discussed circumstances where we assume access to a behavior dataset sampled from $d^{\pi^\mu}_{T_t}(s)$, while being agnostic to its optimality. While it is indeed a practical setting, we also want to evaluate Algorithm 2 when we are provided only expert demonstrations. In such setting, we can still apply results Theorem 4 without loss of its generality, but as a result some of its terms will depend on different distributions and we cannot guarantee a linear dependence on the $M$. Namely, $\epsilon_T$ would instead depend on the number of expert demonstrations, and $C := \sup_{\pi \in \Pi, s \in S, a \in A}\frac{\nu^\pi_{T_l}(s,a)}{p^{\pi^*}_{T_t}(s,a)}$, where they both scale in $O(\frac{1}{(1-\gamma)^2})$. 

Observing the newly defined $C$, we suspect the algorithm's performance would depend on the learner's ability to generate states that is covered by $p^{\pi^*}_{T_t}(s,a)$.
However, recall that Algorithm 2 minimizes the estimated uncertainty of $T_l$, which is trained with samples $s^*, a^* \sim p^{\pi^*}_{T_t}(s,a)$. In other words, $T_l$ admits less uncertainty on $p^{\pi^*}_{T_t}(s,a)$. One could then argue that in such case, by minimizing $T_l$'s uncertainty, $\hat{\pi}$ could learn to match its visitation density with $p^{\pi^*}_{T_t}(s,a)$. Additionally, as alluded in section A, the function $f$ could capture distributional information of states that $\hat{\pi}$ can utilise. 
\\\\
We evaluate this intuition and present its results on Table 3, where the scores are presented in the column labeled $Algorithm2_{expert}$. Indeed, we see that even if $T_l$ is trained solely with expert samples, Algorithm 2 outperforms BC. In the paragraph above, we attibute this to providing the learner with \textbf{more information} that naive BC algorithms fails to utilise. This implies that usage of a estimated model in imitation learning in general could be beneficial.

\begin{figure*}[!t]
\vspace*{-2cm}
    \begin{subfigure}[b]{0.24\textwidth}
        \includegraphics[width=\textwidth]{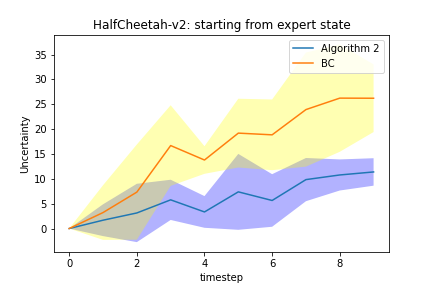}
    \end{subfigure}~
    \begin{subfigure}[b]{0.24\textwidth}
        \includegraphics[width=\textwidth]{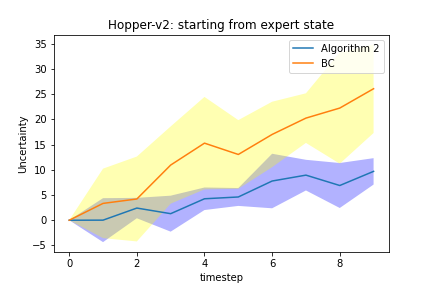}
    \end{subfigure}~
    \begin{subfigure}[b]{0.24\textwidth}
        \includegraphics[width=\textwidth]{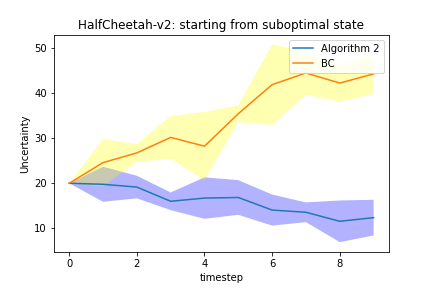}
    \end{subfigure}~
    \begin{subfigure}[b]{0.24\textwidth}
        \includegraphics[width=\textwidth]{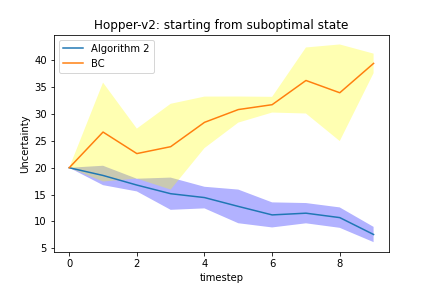}
    \end{subfigure}
    \caption{
    Average uncertainty measurements as a function of timesteps rolled out in the model. where the shaded area represents the std of 5 random seed trials. Uncertainty values are scaled such that states from expert demonstrations obtains 0. 
    }
    \label{fig:MuJoCo}
\end{figure*}

\begin{table}
\label{t3}
\begin{tabular}{c|cccc}
\noalign{\smallskip}\noalign{\smallskip}\hline\hline
Algorithms & $BC$ & Algorithm 2_expert & Algorithm 2_w  \\
\hline
Hopper-v2 & 0.14 $\pm$ 0.18 & 0.68 $\pm$ 0.05  & 0.78 $\pm$ 0.11  \\
HalfCheetah-v2 & 0.55 $\pm$ 0.21 & 0.60 $\pm$ 0.05 &  0.72 $\pm$ 0.11  \\
Walker2d-v2 & 0.10 $\pm$ 0.16 & 0.71 $\pm$ 0.14 & 0.73 $\pm$ 0.22  \\
\hline
\hline
\end{tabular}
 \begin{tablenotes}
      \small
      \item Table 3: Performance evaluations of Algorithm2_expert compared to BC, where for all experiments 3 expert trajectories are provided. 
    \end{tablenotes}
\end{table}

\begin{thebibliography}{9}
\bibliographystyle{unsrtnat}

\bibitem{kiante}Kianté Brantley, Wen Sun, and Mikael Henaff. Disagreement-regularized imitation learning. In \textit{International Conference on Learning Representations}, 2020.
\bibitem{openai}Greg Brockman, Vicki Cheung, Ludwig Pettersson, Jonas Schneider, John Schulman, Jie Tang, and Wojciech Zaremba. Openai gym, 2016.

\bibitem{trex}Daniel Brown, Wonjoon Goo, Prabhat Nagarajan, and Scott Niekum. Extrapolating beyond sub- optimal demonstrations via inverse reinforcement learning from observation. In \textit{International Conference on Machine Learning}, 2019

\bibitem{gcl}Chelsea Finn, Sergey Levine, and Pieter Abbeel. Guided cost learning: Deep inverse optimal control via policy optimization. In \textit{International Conference on Machine Learning}, 2016.

\bibitem{airl}Justin Fu, Katie Luo, and Sergey Levine. Learning robust rewards with adversarial inverse reinforce- ment learning. \textit{International Conference on Learning Representations}, 2018

\bibitem{divergence}Seyed Kamyar Seyed Ghasemipour, Richard Zemel, and Shixiang Gu. A divergence minimization perspective on imitation learning methods. \textit{Conference on Robot Learning}, 2019

\bibitem{imidiv}Liyiming Ke, Sanjiban Choudhury, Matt Barnes, Wen Sun, Gilwoo Lee, and Siddhartha Srinivasa. \textit{Imitation Learning as f-Divergence Minimization} In \textit{arXiv:1905.12888}, 2019



\bibitem{sac}Tuomas Haarnoja, Aurick Zhou, Pieter Abbeel, and Sergey Levine. Soft actor-critic: Off-policy maximum entropy deep reinforcement learning with a stochastic actor. In \textit{International Conference on Machine Learning}, 2018a.

\bibitem{gail}Jonathan Ho and Stefano Ermon. Generative adversarial imitation learning. In \textit{Advances in Neural Information Processing Systems}, 2016.

\bibitem{sail} Fangchen Liu, Zhan Ling, Tongzhou Mu, and Hao Su. State alignment-based imitation learning. \textit{International Conference on Learning Representations}, 2020.

\bibitem{ng}Andrew Y Ng, Stuart J Russell, et al. Algorithms for inverse reinforcement learning. In \textit{International Conference on Machine Learning}, 2000.

\bibitem{firstbc} Dean A Pomerleau. Efficient training of artificial neural networks for autonomous navigation.\textit{ Neural computation}, 1991.

\bibitem{sqil}
Siddharth Reddy, Anca D Dragan, and Sergey Levine. Sqil: imitation learning via regularized behavioral cloning. \textit{International Conference on Learning Representations}, 2020.

\bibitem{reductions}Stéphane Ross and Drew Bagnell. Efficient reductions for imitation learning. In \textit{International Conference on Artificial Intelligence and Statistics}, 2010.

\bibitem{sail2} Yannick Schroecker and Charles L Isbell. State aware imitation learning. In Advances in Neural Information Processing Systems, 2017
\bibitem{maxent}
Brian D Ziebart, Andrew L Maas, J Andrew Bagnell, and Anind K Dey. Maximum entropy inverse reinforcement learning. In AAAI Conference on Artificial Intelligence, 2008.

\bibitem{combo}Tianhe Yu, Aviral Kumar, Rafael Rafailov, Aravind Rajeswaran, Sergey Levine, and Chelsea Finn. COMBO: Conservative offline model-based policy optimization.\textit{ arXiv preprint arXiv:2102.08363}, 2021.

\bibitem{mopo}Tianhe Yu, Garrett Thomas, Lantao Yu, Stefano Ermon, James Zou, Sergey Levine, Chelsea Finn, and Tengyu Ma. MOPO: Model-based offline policy optimization. \textit{arXiv preprint arXiv:2005.13239, 2020}

\bibitem{errorbound} Tian Xu, Ziniu Li, and Yang Yu. Error bounds of imitating policies and environments. \textit{Advances in Neural Information Processing Systems}, 33, 2020.

\bibitem{imilimits}Nived Rajaraman, Lin F Yang, Jiantao Jiao, and Kannan Ramachandran. Toward the fundamental limits of imitation learning. \textit{Advances in Neural Information Processing Systems}, 2020

\bibitem{provablybatch}Yao Liu, Adith Swaminathan, Alekh Agarwal, and Emma Brunskill. Provably good batch reinforce- ment learning without great exploration. \textit{arXiv preprint arXiv:2007.08202}, 2020.

\bibitem{offlinetutorial}Sergey Levine, Aviral Kumar, George Tucker, and Justin Fu. Offline reinforcement learning: Tuto- rial, review, and perspectives on open problems.\textit{ arXiv preprint arXiv:2005.01643, 2020.}

\bibitem{morel}Rahul Kidambi, Aravind Rajeswaran, Praneeth Netrapalli, and Thorsten Joachims. MOReL: Model- based offline reinforcement learning. \textit{arXiv preprint arXiv:2005.05951, 2020}.

\bibitem{mbpo}Michael Janner, Justin Fu, Marvin Zhang, and Sergey Levine. When to trust your model: Model-based policy optimization. \textit{In Advances in Neural Information Processing Systems, pages 12498–12509, }2019.

\bibitem{pessimportance}Jacob Buckman, Carles Gelada, and Marc G Bellemare. The importance of pessimism in fixed- dataset policy optimization. arXiv preprint arXiv:2009.06799, 2020.

\bibitem{noisybc}Fumihiro Sasaki, Ryota Yamashina.
Behavioral Cloning from Noisy Demonstrations. In \textit{International Conference on Learning Representations}, 2021

\bibitem{suboptimal}Yueh-Hua Wu, Nontawat Charoenphakdee, Han Bao, Voot Tangkaratt, and Masashi Sugiyama. Im- itation learning from imperfect demonstration. \textit{arXiv preprint arXiv:1901.09387}, 2019.

\bibitem{mujoco}Emanuel Todorov, Tom Erez, and Yuval Tassa. Mujoco: A physics engine for model-based control. In 2012 \textit{IEEE/RSJ International Conference on Intelligent Robots and Systems}, pp. 5026–5033. IEEE, 2012.

\bibitem{sampleefficient}Fumihiro Sasaki, Tetsuya Yohira, and Atsuo Kawaguchi. Sample efficient imitation learning for continuous control. In \textit{International Conference on Learning Representations}, 2018.

\bibitem{failed}Daniel H Grollman and Aude G Billard. Robot learning from failed demonstrations. \textit{International Journal of Social Robotics}, 4(4):331–342, 2012.

\bibitem{fail}WenSun, AnirudhVemula, ByronBoots,and JAndrewBagnell. Provablyefficient imitation learning from observation alone. In \textit{International Conference of Machine Learning,} 2019.

\bibitem{gan}Ian Goodfellow, Jean Pouget-Abadie, Mehdi Mirza, Bing Xu, David Warde-Farley, Sherjil Ozair, Aaron Courville, and Yoshua Bengio. Generative adversarial nets. In
\textit{Advances in neural information processing systems,}pages 2672–2680, 2014.

\bibitem{gametheoretic}Umar Syed and Robert E Schapire. A game-theoretic approach to apprenticeship
learning. \textit{In Advances in neural information processing systems}, 2008.

\bibitem{rltheory}Alekh Agarwal Nan Jiang Sham M. Kakade Wen Sun.\textit{ Reinforcement Learning: Theory and Algorithms}, 2020

\bibitem{strictlybatch} Daniel Jarrett, Ioana Bica, Mihaela van der Schaar. Strictly Batch Imitation Learning by Energy-based Distribution Matching, In \textit{Advances in neural information processing systems,} 2020

\bibitem{scalable}
Alex J. Chan and Mihaela van der Schaar.
Scalable Bayesian Inverse Reinforcement Learning, 
\textit{International Conference of Learning Representations} 2021

\bibitem{offpolicy}Ilya Kostrikov, Ofir Nachum, Jonathan Tompson. Imitation Learning Via Off-Policy Distribution Matching. \textit{International Conference of Learning Representations} 2020

\bibitem{primal} Robert Dadashi, Léonard Hussenot, Matthieu Geist, Olivier Pietquin. Primal Wassertein Imitation Learning. \textit{International Conference of Learning Representations} 2021

\bibitem{dagger}Stéphane Ross Geoffrey J. Gordon J. Andrew Bagnell. A Reduction of Imitation Learning and Structured Prediction
to No-Regret Online Learning. In \textit{AISTATS}, 2011

\bibitem{modelbased}Peter Englert; Alexandros Paraschos; Jan Peters; Marc Peter Deisenroth.  Model-based imitation learning by probabilistic trajectory matching.  2013 \textit{IEEE International Conference on Robotics and Automation}

\bibitem{bclimitation}Felipe Codevilla, Eder Santana, Antonio M. López, Adrien Gaidon. Exploring the Limitations of Behavior Cloning for Autonomous Driving. In \textit{	arXiv:1904.08980}

\bibitem{ppo}John Schulman, Filip Wolski, Prafulla Dhariwal, Alec Radford, and Oleg Klimov. Proximal policy optimization algorithms. \textit{arXiv preprint arXiv:1707.06347, 2017}.

\bibitem{agarwal}Alekh Agarwal, Sham Kakade, Akshay Krishnamurthy, and Wen Sun. Flambe: Structural complexity and representa- tion learning of low rank mdps. \textit{NeurIPS}, 2020b.


\end{thebibliography}
\end{document}